\documentclass[a4paper,conference]{IEEEtran}
\usepackage{times}
\usepackage{epsfig}
\usepackage{graphicx}
\usepackage{amsmath}
\usepackage{amssymb}
\usepackage{subcaption}
\usepackage{wrapfig}
\usepackage[utf8]{inputenc}
\usepackage[T1]{fontenc}
\usepackage{style}
\usepackage{makecell}
\usepackage{graphicx}
\usepackage{url}
\usepackage{multirow}
\usepackage{xcolor}
\usepackage{float}
\usepackage{booktabs}
\usepackage{caption} 

\begin{document}

\title{\LARGE \bf
VidHarm: A Clip Based Dataset for Harmful Content Detection 
}
\author{Johan Edstedt$^1$ \quad Amanda Berg$^1$\quad Michael Felsberg$^1$ \and\quad Johan Karlsson$^2$\quad Francisca Benavente$^2$\quad 
 Anette Novak$^2$ \and\quad Gustav Grund Pihlgren $^3$
 \\
 $^1$Computer Vision Laboratory, Linköping University \quad $^2$Statens Medieråd \and\quad $^3$EISLAB Machine Learning, Luleå University of Technology
 \\
 $^1${\tt\small{\{firstname\}.\{lastname\}@liu.se}}
 \quad
 $^2${\tt\small{\{firstname\}.\{lastname\}@statensmedierad.se}}
 \and\quad
 $^3${\tt\small{\{firstname\}.\{lastname\}@ltu.se}}
 }
 
\author{\IEEEauthorblockN{Johan Edstedt}
\IEEEauthorblockA{Computer Vision Laboratory\\
Linköping University}
\and\quad
\IEEEauthorblockN{Amanda Berg}
\IEEEauthorblockA{Computer Vision Laboratory\\
Linköping University}
\and\quad
\IEEEauthorblockN{Michael Felsberg}
\IEEEauthorblockA{Computer Vision Laboratory\\
Linköping University}
\and\quad
\IEEEauthorblockN{Johan Karlsson}
\IEEEauthorblockA{Statens Medieråd\\
Stockholm}
\and\quad
\IEEEauthorblockN{Francisca Benavente}
\IEEEauthorblockA{Statens Medieråd\\
Stockholm}
\and\quad
\IEEEauthorblockN{Anette Novak}
\IEEEauthorblockA{Statens Medieråd\\
Stockholm}
\and\quad
\IEEEauthorblockN{Gustav Grund Pihlgren}
\IEEEauthorblockA{EISLAB Machine Learning\\
Luleå University of Technology}}

\author{
\IEEEauthorblockN{Johan Edstedt\IEEEauthorrefmark{1},
Amanda Berg\IEEEauthorrefmark{1},
Michael Felsberg\IEEEauthorrefmark{1}\\
Johan Karlsson\IEEEauthorrefmark{2},
Francisca Benavente\IEEEauthorrefmark{2},
Anette Novak\IEEEauthorrefmark{2}\\
Gustav Grund Pihlgren\IEEEauthorrefmark{3}}
\IEEEauthorblockA{\IEEEauthorrefmark{1}Computer Vision Laboratory, Linköping University}
\IEEEauthorblockA{\IEEEauthorrefmark{2}Statens Medieråd, Stockholm}
\IEEEauthorblockA{\IEEEauthorrefmark{3}EISLAB Machine Learning, Luleå University of Technology}
}

\maketitle


\begin{abstract}

Automatically identifying harmful content in video is an
important task with a wide range of applications. However,
there is a lack of professionally labeled open datasets
available. In this work VidHarm, an open dataset of 3589
video clips from film trailers annotated by professionals,
is presented. An analysis of the dataset is performed,
revealing among other things the relation between clip and
trailer level annotations. Audiovisual models are trained
on the dataset and an in-depth study of modeling choices
conducted. The results show that performance is greatly
improved by combining the visual and audio modality,
pre-training on large-scale video recognition datasets, and
class balanced sampling. Lastly, biases of the trained
models are investigated using discrimination probing.

VidHarm is openly available, and further details are
available at the webpage \url{https://vidharm.github.io/}



\end{abstract}

\section{Introduction}

Identifying harmful content in motion picture is a well-established field. Motion picture rating systems are used in many countries, with the most well-known being the Motion Picture Association (MPA) system in the United States. In all these systems, to the best of our knowledge, ratings are set by human expertise alone. In practice, this means that one or more professional classifiers make a joint decision for each movie, based on a (perhaps loose) set of guidelines. What constitutes harmful content may include factors such as violence, sexual content, and horror.

Automatic detection and classification/rating of harmful content has, however, so far mostly been directed at detecting specific forms of harmful  content in videos.
Such approaches fall short in cases outside their area of detection and may fail to provide nuance to their predictions.
There is likely a multitude of reasons for this.
MPA and other ratings are typically provided for feature film, and this content is typically under strict copyright laws which prevent open-source datasets for the public.
Further complications arise from the extreme sparsity of the labels, for a typical film of 1.5 hours there are approximately $10^5$ frames of content.
This makes modern end-to-end methods impossible to train in practice and requires rough simplifications, e.g., by only considering dialogue. While such simplifications may yield good results, they fail to capture a general understanding of what constitutes harmfulness.
Therefore, a fine-grained dataset for harmful content detection is sorely needed.

For these reasons, this work introduces a high-quality video dataset based on 3589 video clips of length 10 seconds with labels based on an established film classification system, labeled by professional film classifiers.
Furthermore, a thorough study of the dataset has been performed, including baseline experiments for age rating classification using video-only, audio-only, and combined audio and video modalities.

The contributions of this work are as follows:
\begin{enumerate}
    \item VidHarm, a diverse open-source professionally labeled fine-grained dataset for harmful content detection based on an established film classification system, is provided.
    \item A thorough analysis of VidHarm, including establishing connections between clip and trailer ratings, and an investigation of potential biases.
    \item Studies of video recognition models utilizing both modalities of VidHarm is performed. Additionally, a task-appropriate evaluation metric is proposed, and compared to previous metrics.
\end{enumerate}

\begin{table*}[h!]
    \centering
    \caption{
        Comparison of the proposed and previous datasets. $^\dag$Labels and parts of data is openly available, while other parts must be purchased. $^*$Potentially occational phrases or short conversations in additional languages.
    }

    \begin{tabular}{c|c|c|c|c|c|c|c|c|c}
    \toprule
         Datasets &  Video & Text & Open & Harmful Content & Violence & Affective Impact & Expert Labels & Age Categories & Languages \\
        \midrule
        \cite{Shafaei2020} \& \cite{mohamed2020first} & $\times$ & $\checkmark$ & $\checkmark$ & $\checkmark$ & $\checkmark$ & $\times$ & $\checkmark$ & 5 \& 5/6 & 1$^*$ \& 1$^*$\\
        Papadamou~\cite{Papadamou2019} & $\checkmark$ & $\times$ & $\checkmark$ & $\checkmark$ & $\checkmark$ & $\times$ & $\times$ & 4 & 1$^*$\\
        VSD~\cite{schedi2015vsd2014} & $\checkmark$ & $\times$ & $\times^{\dag}$ & $\times$ & $\checkmark$ & $\checkmark$ & $\times$ & 0 & 1$^*$\\
        LIRIS-ACCEDE~\cite{baveye2015liris} & $\checkmark$ & $\checkmark$ & $\checkmark$ & $\times$ & $\times$ & $\checkmark$ & $\times$ & 0 & 1$^*$\\
        VidHarm (Proposed) & $\checkmark$ & $\times$ & $\checkmark$ & $\checkmark$ & $\checkmark$ & $\times$ & $\checkmark$ & 4 & 37$^*$\\
        \bottomrule
    \end{tabular}
    \label{tab:datasets}
\end{table*}

\section{Related Work}
\parsection{Defining Harmfulness}
Is violence harmful? Is nudity harmful? Is horror harmful? And to what degree should different factors impact the rating? Such judgements are difficult to make for a layperson.
To this end, the Swedish film classification system is adapted as the base for the dataset and labeling has been conducted by professional classifiers.

The Swedish film classification system is set with the well-being of children as the main goal and is regulated under Swedish law\footnote{SFS 2018:1964, SFS 2010:1882}. The term well-being is the same that is used in UN:s Convention on the Rights of the Child \cite{assembly1989convention}, which states that children have the right to take part of information, as well as a right to be protected from content that can be harmful to their well-being. %

Harmful content is then defined as content that may elicit anxiety, fear, or other detrimental effects in children. The ratings, however, are not recommendations and do not reflect whether the film is appropriate or suitable for a certain age group. Religious, political, or moral attitudes are not taken into account when deciding the age ratings \cite{filmclassification}. To be able to determine these ratings, Swedish film classifiers typically have backgrounds in child psychology or similar fields, and screenings of films are regularly shown to children to gain a better understanding of what constitutes harmful content.

The system does not have a fixed set of instructions from which an objective rating can be set.
This means that the rating cannot be set by a novice.
In comparison to the typically straightforward classification tasks in computer vision, this makes the labeling setting more complex and challenging.
Furthermore, the chosen system specifically classifies harmfulness, and not other factors such as appropriateness which is taken into account in, e.g., MPA ratings \cite{mpaa_ratings}. 

While no exact guidelines exist for classifying harmfulness in the Swedish system, an approximate guideline is provided to give an intuition for the reasoning behind different ratings and can be found in Table \ref{tab:guideline}.
Note that these guidelines are reductive and do not cover the full spectrum of harmful content. 
For nuance, the work of Andersson \etal \cite{Andersson2016} provides a more complete discussion.
\newcolumntype{L}{>{\arraybackslash}m{0.76\columnwidth}}

\begin{table}[]
    \centering
    \caption{An overview of the Swedish rating system. The system is comprised of 4 ratings, BT (all ages), 7+ (restricted to 7 years and older), 11+ (restricted to 11 years and older), and 15+ (restricted to 15 years and older).}
    \begin{tabular}{l|L}
    \toprule
    Rating & Approximate guideline \\
    \midrule
BT& Does not contain any seriously frightening content. Events such as frightening noises and high-paced shot-transitions may be sufficient to warrant a higher rating.
May contain light-hearted thrilling scenes, and playful violence.\\
\midrule
7+& May contain scenes with mild action elements such as fighting and shooting, as well as mildly frightening scenes and scary effects in, e.g., cartoons.
\\
\midrule
11+& A clear emphasis is placed on realism and personal identification with the content. Less realistic scenes with more pronounced violence may receive a lower rating than a more realistic and personally relatable counterpart. \\
\midrule
15+& Detailed violence, large amounts of blood, and suffering are typically factoring in determining this rating. Highly threatening and strongly anxiety inducing scenes as well as depictions of sexual abuse also belong to this rating.\\
\bottomrule

    \end{tabular}

    \label{tab:guideline}
\end{table}

\parsection{Detecting Harmful Content in Video} 
Early work in harmful content detection focused on handcrafted ways of identifying violence or other harmful specific types of harmful content in video. Shot activity and shot length~\cite{Vasconcelos1997}, flame detection, blood detection, and burst sounds~\cite{Nam1998} could be used. Later works focused on handcrafted feature extraction combined with support vector machines~\cite{Acar2013,Giannakopoulos2006,Wang2011}, or Bayesian networks~\cite{Giannakopoulos2010}. Recent works~\cite{Dai2014,Dai2015,Martinez2019,Wehrmann2018,Singh2019,martinez2020joint}, incorporate deep learning methods and move away from handcrafted features. These methods are all restricted in the sense that they only detect certain forms of harmful content, most often violence.
In contrast to these works, this work is not restricted to a specific form of harmful content.

Directly classifying harmful content with respect to an established rating system has, interestingly and somewhat surprisingly, received far less attention from the research community. In a recent work, Shafaei \etal~\cite{Shafaei2020} predicts the MPA rating for movies based on the movie script. Mohamed and Ha~\cite{mohamed2020first} similarly predict the MPA and British Board of Film Classification (BBFC) ratings based on transcribed dialogue. While they show impressive results, it could be argued that ignoring the visual modality as well as environmental and musical elements of the soundtrack, hampers a more general understanding of what is harmful.
In contrast to these works, this work makes full use of both the visual and audio modality. 


\parsection{Previous Datasets}
VSD~\cite{schedi2015vsd2014,constantin2020affect} is a dataset aimed at classification of violence, and \emph{affective impact} (i.e. the emotional impact) of movie scenes. However, due to the copyrighted material, they are unable to provide the data openly, and researchers must purchase the films to gain access. Furthermore, their classification of violent content is not done by experts. Compared to that work, the proposed dataset is openly available, classifies a broader range of harmful content, and is labeled by experts. The LIRIS-ACCEDE~\cite{baveye2015liris} dataset is a clip-based dataset with focus on affective impact. They provide the data openly, since they use less restricted material. However, they do not provide any harmful content classification.

Papadamou \etal\cite{Papadamou2019} construct a dataset with MPA-inspired labels from YouTube videos to identify inappropriate content. However, they restrict their model to a set of handpicked features from the video. Additionally their labeling process is not performed by experts.

Both Shafaei \etal~\cite{Shafaei2020} and Mohamed and Ha~\cite{mohamed2020first} provide datasets for the MPA and British Board of Film Classification (BBFC) ratings based on transcribed dialogue. However, these datasets do not take non-verbal audio nor visual content into account. A summary of the relationships between the proposed dataset and already existing ones is shown in Table~\ref{tab:datasets}.
\section{Dataset}
\label{sec:dataset}
\subsection{Data Acquisition}
\label{sec:data_acquisition}
VidHarm consists of over 300 recently released film trailers which was acquired through two Swedish film distributors.
The trailers were manually gathered, and checked for duplicates, to ensure the quality of the data.
All trailers were downloaded in at least 720p quality.
A very simple clipping process is used where each trailer is divided into 10-second disjointed clips.
These clips are given hash names and are then given in random order to two professional film classifiers. Both have extensive experience working for the Swedish Media Council, one with 15 years in the job, the other with 12 years. One of annotators in this work is a licensed psychologist with a Master of Science in Psychology, the other a licenced school-teacher (compulsory school) with 20 years experience of teaching in a public school 3rd to 6th grade (9-12 years) and responsible for child right issues at the Swedish Media Council. Both have perennial experience of working with children. We argue that this background is important when producing annotations, as it may be difficult for a lay person to judge what may or may not be harmful to children.

When a clip has been given two reviews, a check for agreement is made. For all clips where the annotators disagree, a second review is conducted jointly and a refined label produced.
VidHarm is labeled twice for multiple reasons.
The main reason is to ensure that the dataset is consistent and unbiased.
This is important since it may be difficult for a layperson to judge whether the labels are accurate or not.
Additionally, this closely corresponds to the procedure used to classify films for the Swedish market, where two classifiers are required by law to make a joint decision.
In this way, the data collection stays as close as possible to the working process that the classifiers are used to.

Furthermore, the additional label enables refinement of the dataset by relabelling all samples with disagreement.
An alternative method would be to train models on the dataset and reassess all labels which the model and label disagree \cite{Beyer2020,northcutt2021pervasive}.
However, this method requires the trained models to be extremely accurate which is difficult to achieve, especially early in the data collection process.

In addition to providing the original clips with labels, a converted version where each clip is converted into a set of images, an audio file, and a precomputed log mel-scaled \cite{Stevens1937} spectrogram is provided.

Finally, the labels for the entire trailers to which the clips correspond are also provided. Due to being relatively fewer, the labeling was done jointly by both labelers for all trailers.

\subsection{Data Description}
\label{sec:data_description}
\begin{figure*}
     \begin{subfigure}[b]{.3\linewidth}
    \includegraphics[width=\linewidth]{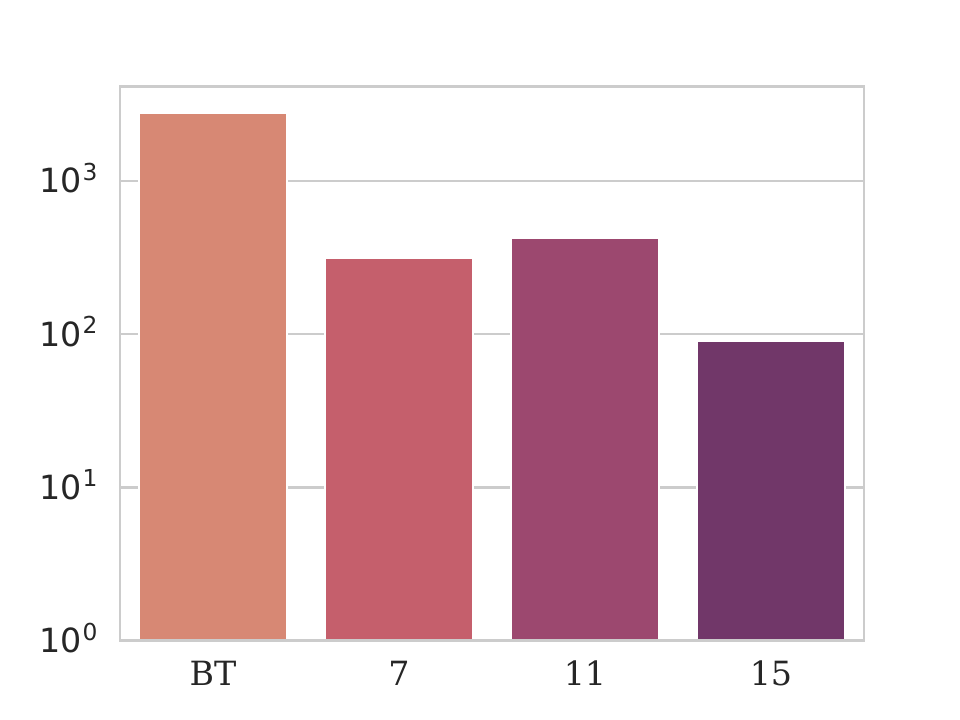}
    \caption{Rating distribution of clips in our dataset.}
    \label{fig:clip_classdistribution}
    \end{subfigure}
    \begin{subfigure}[b]{.3\linewidth}
    \includegraphics[width=\linewidth]{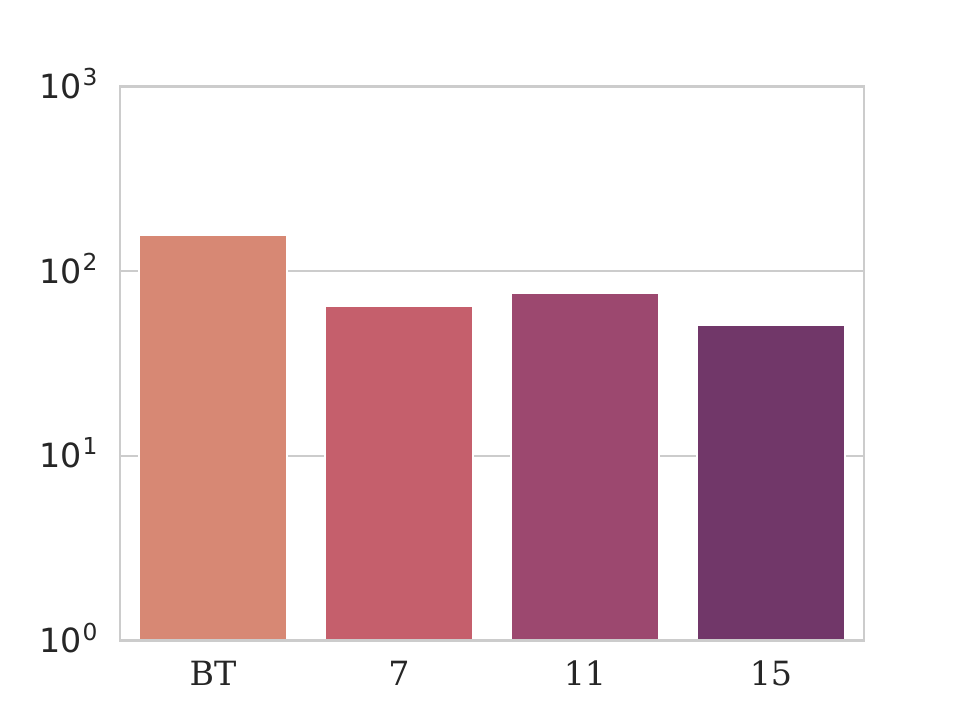}
    \caption{Rating distribution for full trailers in our dataset.}
    \label{fig:trailer_classdistribution}    
    \end{subfigure}
    \begin{subfigure}[b]{.3\linewidth}
    \includegraphics[width=\linewidth]{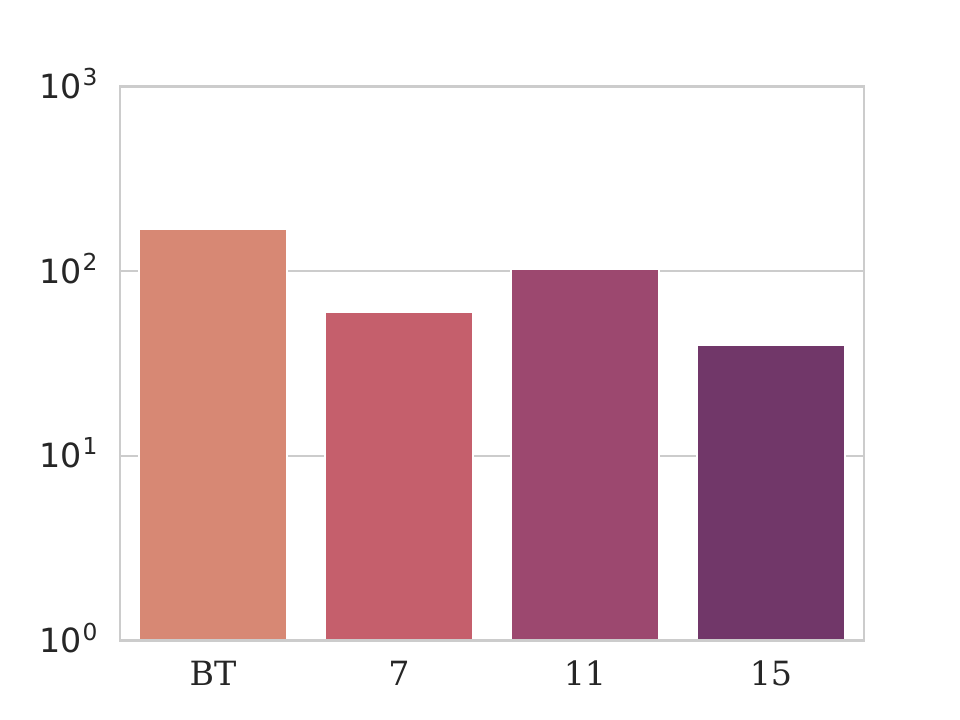}
    \caption{Rating distribution for full trailers in the Swedish Market (2019).}
    \label{fig:comp_classdistribution}    
    \end{subfigure}
    \caption{Rating distribution of our created dataset compared to the distribution in the Swedish market for trailers. Height represents the number of clips/trailers with a certain rating.}
    \label{fig:classdistribution}
\end{figure*}

The dataset is composed of 3589 non-overlapping 10 second clips taken from 351 film trailers. Film trailers were chosen for multiple reasons. One such reason reason is that it is typically illegal to distribute full feature films without holding rights, and acquiring rights is prohibitively expensive and/or difficult. Trailers are not subject to such harsh copyright restrictions, and by an agreement with two film distributors we are able to provide free access to their trailers for research purposes. Trailers do additionally have multiple features that make them suitable for this works purposes. Trailers have less frame-to-frame correlation than full feature film since shots in trailers are typically much shorter than the full-length counterparts. This allows a richer dataset to be provided without having to use massive amounts of data, and without having to resort to complicated content mining. Furthermore, trailers tend to have a higher frequency of dramatic scenes, this eases some issues with unbalanced data since the data would otherwise risk being extremely unbalanced toward the BT class. The distribution of ratings in the dataset is presented in Fig.~\ref{fig:classdistribution}. Notably, it can be seen that the full trailers from which the clips are sampled from, closely resembles the Swedish market for trailers, with a skew towards more 15 ratings. This skew is likely due to a selection bias whereby trailers more likely to contain harmful content were selected, to get a greater diversity in the rating distribution.
The dataset is multi-lingual, with an impressive 37 different spoken languages. The most common languages used being English, French, Swedish, Spanish, and German.

\subsection{Evaluation}
The data is partitioned into training, validation, and test sets with 60\%, 10\%, and 30\% of the total data respectively. These partitions are constructed in such a way that all clips from the same trailer are assigned to the same partition. This ensures that there is no data leakage from the test set to the training set.

Age classification makes use of ordinal labels and choosing a metric that takes into account the distance between labels is more appropriate than distance agnostic metrics such as accuracy. We propose the \emph{Class Absolute Error} (CAE) as our main evaluation metric, which can be defined as
\begin{equation}
\text{CAE}_{\ell}(p_{\theta},D)=\frac{\sum_{(x,y_{\ell})\in D_{\ell}}|y_{\ell}-\mathbf{E}_{y\sim p_{\theta}(y|x)}\left[y\right]|}{|D_{\ell}|},
\end{equation}
where $D$ is the dataset, $D_{\ell}$ is a subset containing only the video-label pairs with label $\ell$, $y_{\ell}$ is the ordinal value of the age-rating given the video clip $x$, and $p_{\theta}(y|x)$ is the model prediction given $x$. The BT label is mapped to the ordinal value 3, and 7, 11, and 15 to their natural ordinal values. Note that this definition of the metric is applicable both to regressive and classifier models. 
The proposed metric has the benefit of being label balanced while retaining their inherent geometric relations. For example, if CAE(7+) = 2, it can be interpreted as that the average error is 2 years whenever the ground truth label is 7+. The metric is the multi-class generalization of the \emph{MAE} metric, commonly used in, e.g., age estimation \cite{lanitis2004comparing}, where the class-wise distinction has also sometimes been used \cite{geng2007automatic}. The metric may be further simplified into the \emph{mean Class Absolute Error} (mCAE) by taking the mean over all the classes. This makes model comparison simpler while sacrificing some of the granularity of the CAE metric. If the dataset is class balanced mCAE is equivalent to MAE.

To complement this metric, the multi-class mean Precision (mP) and mean Recall (mR) metrics were also used. These are defined as
\begin{equation}
\label{eq:mp}
    \text{mP} = \frac{1}{|L|}\sum_{\ell\in L} \frac{\text{TP}_{\ell}}{\text{TP}_{\ell} + \text{FP}_{\ell}},
\end{equation}
\begin{equation}
\label{eq:mr}
    \text{mR} = \frac{1}{|L|}\sum_{\ell\in L} \frac{\text{TP}_{\ell}}{\text{TP}_{\ell} + \text{FN}_{\ell}}.
\end{equation}
\subsection{Data Analysis}
For all datasets, an estimate of the labeling error is vital. In Table \ref{tab:humanlabel} an estimate of the individual annotation error in the data is presented. This is a reasonable human-level performance indicator and performing better than this could be considered superhuman performance. Note that the BT label is, for humans, the simplest to label correctly. This is possibly due to the fact that it is often marked by the absence of harmful content, while the other labels must distinguish between different levels of harmfulness. However, models trained on VidHarm did not in general perform better on the BT label than the other labels.
\begin{table}[]
    \centering
        \caption{Individual label errors, acquired by comparing the individual human classifiers to their jointly refined predictions. Measured in CAE (lower is better).}

    \begin{tabular}{l|ccccc}
    \toprule
    &\multicolumn{5}{c}{CAE~$\downarrow$} \\

         &BT  &  7+ & 11+ & 15+ & Mean \\
         \midrule

         Individual Classifier &  0.04 & 1.00 & 1.13 & 0.99 &	0.79 
         \\
    \bottomrule
    \end{tabular}
    \label{tab:humanlabel}
\end{table} 

An interesting question is the relation between the clip-level labels and the trailer-level labels. For this purpose a simple model is proposed. The model simply predicts the highest rated clip as the rating of the trailer in full. It turns out that this is a quite powerful predictor, as demonstrated in Table \ref{tab:clip_to_trailer}. \begin{wrapfigure}{r}{0.25\textwidth}
    \centering
    \includegraphics[trim=15 0 30 15,clip,width=0.25\textwidth]{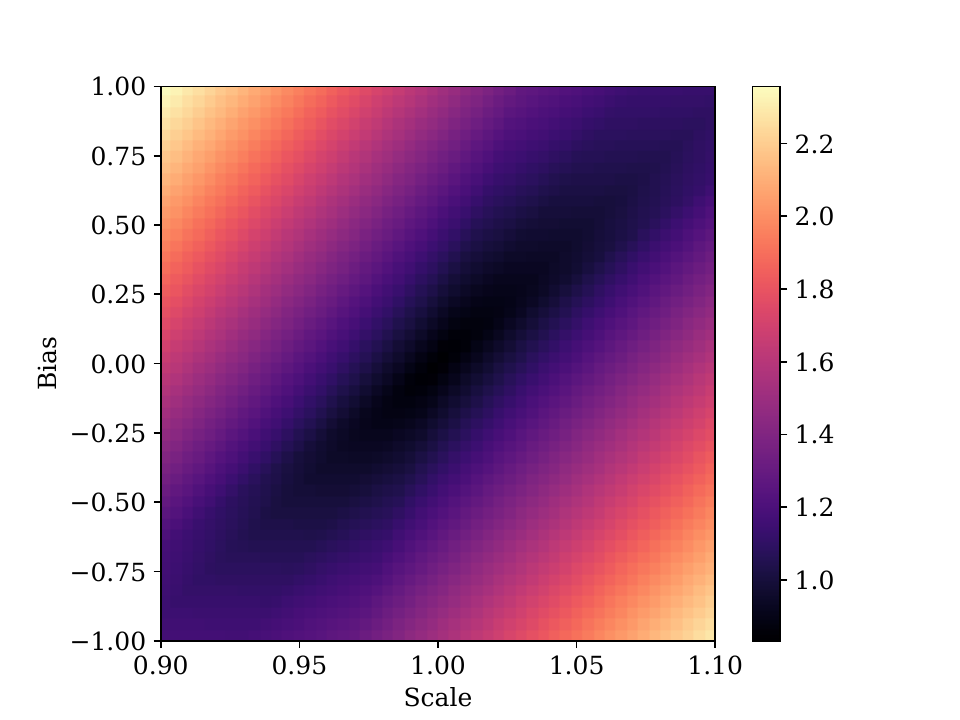}
    \caption{Highest rated clip vs rating of full trailer. Measured in mAE (lower is better).}
    \label{fig:clip_to_trailer_grid}
\end{wrapfigure} Furthermore, a grid search over the space of affine combinations shows that the identity predictor is close to, if not the best possible affine predictor, as can be seen in Figure \ref{fig:clip_to_trailer_grid}. The signed error residual is furthermore close to unbiased, with $\mathbf{E}(r) = -0.12$.  This seems to indicate that the clip level annotations are fundamentally connected to the full trailer annotations. While this is perhaps not surprising, it has not been investigated previously, and further validates the approach in using short clips.
\begin{table}[]
    \centering
        \caption{Rating error between the highest rated clip in a trailer, and the rating of the trailer in full. Measured in CAE (lower is better)}

    \begin{tabular}{l|ccccc}
    \toprule
        &\multicolumn{5}{c}{CAE~$\downarrow$} \\

         &BT  &  7+ & 11+ & 15+ & Mean \\
         \midrule

         Clip-to-Trailer &  0.38 &1.35 &0.53 & 1.02 & 0.82
         \\
    \bottomrule
    \end{tabular}

    \label{tab:clip_to_trailer}
    \vspace{-12pt}
\end{table}

\subsection{Relevance}
\label{sec:dataset_relevance}
While some similar datasets exists, as shown in Table~\ref{tab:datasets}, there are some key differences that shows the relevance of the proposed dataset.
Most notably there is a complete lack of a dataset with professionally annotated videos for age rating or similar tasks.
Additionally, none of the prior video dataset are annotated by official age ratings.
Rather VSD~\cite{schedi2015vsd2014} has binary violence classifications, LIRIS-ACCEDE~\cite{baveye2015liris} is a regression dataset with two axes, arousal and valence, none of which correspond directly to age ratings, and Papadamou~\cite{Papadamou2019}, which uses four ad-hoc age rating categories.
While the difference between the proposed video dataset and the professionally labeled text-based datasets is obvious, it should be noted that the proposed dataset provide professional labels for individual clips while those datasets use the age rating of the entire movie for each data-point.



\section{Baselines}
This section details the setup for experiments performed on the dataset in order to present baseline results on the data over video-only, audio-only, and combined modalities.

\subsection{Models}
To properly model the data, both audio and visual modalities must be used.
For the visual modality, the recently proposed SlowFast~\cite{Feichtenhofer2019} networks are used.
SlowFast networks use a fast and a slow branch to capture the full temporal dynamics of video, and have proven themselves as efficient but powerful video prediction tools~\cite{zhu2020comprehensive}.
This work uses an 8x8 50-layer model, pretrained on the Kinetics-400 dataset.

For the audio modality, a standard ResNet~\cite{He2016} model with 18 layers pretrained on Kinetics-400 is used.
The input consists of mel-scaled spectrograms, which are the most common feature extraction method for audio~\cite{georgescu2021performance} and have previously been shown to perform well~\cite{Park2019}.
Finally, several different fusions of the unimodal models are tested.

\parsection{Fusing Video and Audio Predictions}
Two simple models of fusing labels will be considered.
The two cases can be found in Eq.~\ref{eq:independent_modes} and Eq.~\ref{eq:fused_modes} respectively, where $p_{\theta}(y|x)$ is the model predictive distribution over ratings $y$ given the video clip $x$.
\vspace{-1mm} 
\begin{equation}
    p_{\theta}(y|x)\propto p_{\theta}(y|x_{\mathrm{visual}})p_{\theta}(y|x_{\mathrm{audio}}),
    \label{eq:independent_modes}
\end{equation}
\vspace{-3mm} 
\begin{equation}
    p_{\theta}(y|x)\propto p_{\theta}(y|x_{\mathrm{visual}})+p_{\theta}(y|x_{\mathrm{audio}}).
    \label{eq:fused_modes}
\end{equation}

The first case with independent predictions is attractive as the information conveyed by the visual and audio modality is of different/independent character.
On the other hand, the case with fused probability space is attractive when the predictions are strongly correlated with some independent zero mean noise.

Furthermore a learned approach will be considered. In this approach 2-stage training is conducted.
First, both modalities are trained independently. 
Then the networks are frozen and a small multi-layer perceptron (MLP) is trained using the predictive distribution of both modalities as input.
While other methods of training have been recently proposed, whereby visual and audio information is fused iteratively inside a single joint network~\cite{kazakos2019epic,wang2020makes,Xiao2020}, such approaches are outside the scope of this work.


\subsection{Training and Testing Details}

\parsection{Video}
The video data is preprocessed by sampling every $7$\textsuperscript{th} frame and grouping $32$ consecutive sampled frames into a datapoint.
During training time the frames are scaled randomly in the range $[3/4, 4/3]$ and then to $224\times 224$ pixels.
During testing the height or each frame is scaled to $256$ pixels, take a center crop of $256\times256$, and use two clips.

\parsection{Audio}
Audio is preprocessed into spectograms by using Hann windowed Fourier transform of 32ms with a 16ms hop length.
The spectograms are converted to are log mel-scaled spectograms with 128 mels in the range 20-15000~Hz which are normalized using the empirical mean and variance of the dataset.
During training the audio is divided into 5 second clips and a modified version of SpecAugment~\cite{Park2019} without time-warping is used.
During testing the entire audio is used without clipping and SpecAgument.

\parsection{General}
Both modalities use cross-entropy as loss function, 16 as batch size, and an SGD optimizer with a momentum of $0.9$ and a learning rate of $10^{-3}$.
All experiments were run using a single RTX 6000 GPU.

\section{Potential Learned Discrimination}

It is well-known that machine learning models often learn to include sensitive attributes in their decision making~\cite{silberg2019notes}.
Even if the people who label the data do not let sensitive attributes affect them, it may be the case that the underlying data is already skewed. In this work, gender imbalances in learned age ratings are investigated.


In order to test trained models for unwanted gender bias, the model is probed by inputting similar scenes where the \emph{only} notable difference is the gender of the central actor of the scene.
If the model systematically rates the scene differently between genders, an undesired bias exists.
To make this probing a subset of the DeepFakeDetection dataset is used which contains 363 videos created by having 28 paid actors act out 16 different scenes~\cite{google2019contributing, roessler2019faceforensicspp}. Note that none of the scenes contain harmful content.
Trained models were then probed by having analyzing the videos and outputting a prediction for the BT class.

\section{Results}
\begin{table*}[h!]
    \caption{Ablation study over the effect of pretraining (PT) and class balanced sampling (CB) on visually based model. Performance measured in CAE (lower is better), mP and mR (higher is better).}

    \centering
    \begin{tabular}{l|ccccccc}    
    \toprule
    &\multicolumn{5}{c}{CAE~$\downarrow$} &mP~\%~$\uparrow$ & mR~\%~$\uparrow$\\
    & BT & 7+ & 11+ & 15+ & Mean & &\\  
\midrule
Baseline Visual no CB/PT&	1.02 $\pm$ 0.3 & 2.38 $\pm$ 0.5 & 6.07 $\pm$ 0.6 & 9.30 $\pm$ 0.9 & 4.69 $\pm$ 0.4 & 25.37 $\pm$ 6.18 & 25.89 $\pm$ 0.99\\

Baseline Visual no CB & 1.11 $\pm$ 0.2 & 1.98 $\pm$ 0.1 & 4.32 $\pm$ 0.2 & 5.77 $\pm$ 0.2 & 3.54 $\pm$ 0.1 & 32.47 $\pm$ 0.50 & 33.46 $\pm$ 0.98\\

Baseline Visual no PT & 4.48 $\pm$ 0.7 & 1.96 $\pm$ 0.8 & 2.28 $\pm$ 0.5 & 4.67 $\pm$ 1.0 & 3.35 $\pm$ 0.1 & 30.08 $\pm$ 2.09 & 37.54 $\pm$ 3.57\\

Baseline Visual  & 2.61 $\pm$ 0.5 & \textbf{1.68} $\pm$ 0.1 & 2.86 $\pm$ 0.4 & \textbf{3.48} $\pm$ 0.6 & \textbf{2.66} $\pm$ 0.1 & \textbf{44.06} $\pm$ 4.37 & \textbf{52.31} $\pm$ 3.21\\
\bottomrule
    \end{tabular}
    \label{tab:pretraining_cb_V}
\end{table*}

\begin{table*}[h!]
    \caption{Ablation study over the effect of pretraining (PT) and class balanced sampling (CB) on audio based models. Performance measured in CAE (lower is better), mP and mR (higher is better).}

    \centering
    \begin{tabular}{l|ccccccc}    
    \toprule
    &\multicolumn{5}{c}{CAE~$\downarrow$} & mP~\%~$\uparrow$ & mR~\%~$\uparrow$\\
    & BT & 7+ & 11+ & 15+ & Mean & & \\  
\midrule
Baseline Audio no CB/PT&	
0.47 $\pm$ 0.4 & 2.99 $\pm$ 0.7 & 6.25 $\pm$ 1.0 & 8.95 $\pm$ 1.5 & 4.66 $\pm$ 0.8 & 31.08 $\pm$ 9.31 & 27.18 $\pm$ 3.50\\

Baseline Audio no CB & 
1.02 $\pm$ 0.2 & 2.10 $\pm$ 0.3 & 3.73 $\pm$ 0.7 & 5.20 $\pm$ 0.6 & 3.01 $\pm$ 0.3 & 31.68 $\pm$ 0.45& 35.89 $\pm$ 3.26\\

Baseline Audio no PT & 
3.34 $\pm$ 1.4 & 2.48 $\pm$ 1.2 & 2.46 $\pm$ 0.6 & 3.52 $\pm$ 1.2 & 2.95 $\pm$ 0.4 & \textbf{41.62} $\pm$ 6.64& 44.31 $\pm$ 5.82\\
Baseline Audio  &  3.21 $\pm$ 0.3 & 2.11 $\pm$ 0.3 & 2.00 $\pm$ 0.2 & 3.47 $\pm$ 0.4 & \textbf{2.70} $\pm$ 0.1 & 38.66 $\pm$ 1.72 & \textbf{44.64} $\pm$ 3.84\\
\bottomrule
    \end{tabular}
    \label{tab:pretraining_cb_A}
\end{table*}
\begin{table*}[h!]
    \caption{Study over the effect of fusion method over visual and audio modalities. Measured in CAE (lower is better), mP and mR (higher is better). Baseline denotes the pretrained models using class balanced sampling.}

    \centering
    \begin{tabular}{l|ccccccc}
    \toprule
    &\multicolumn{5}{c}{CAE~$\downarrow$}  & mP~\%~$\uparrow$ & mR~\%~$\uparrow$\\
    & BT & 7+ & 11+ & 15+ & Mean & &\\
    \midrule
Baseline Fusion - Averaged	& 2.91 $\pm$ 0.4 & 1.53 $\pm$ 0.2 & \textbf{2.21} $\pm$ 0.2 & 3.47 $\pm$ 0.2 & 2.53 $\pm$ 0.1 & 45.12 $\pm$ 1.97 & 52.62 $\pm$ 2.09\\
Baseline Fusion - Independent &	1.96 $\pm$ 0.5 & 1.90 $\pm$ 0.1 & 2.53 $\pm$ 0.2 & 2.71 $\pm$ 0.4 &\textbf{2.29} $\pm$ 0.1 & 46.07 $\pm$ 2.93& 52.91 $\pm$ 2.81\\
Baseline Fusion - MLP & 2.20 $\pm$ 0.1 & 1.63 $\pm$ 0.2 & 2.60 $\pm$ 0.1 & 2.74 $\pm$ 0.2 & \textbf{2.29} $\pm$ 0.1 & \textbf{49.11} $\pm$ 1.93 & \textbf{57.58} $\pm$ 1.81\\
\bottomrule
    \end{tabular}
    \label{tab:fusion}
\end{table*}
In this section, the results of the baseline experiments on the proposed dataset are presented.
All performance numbers are taken as the average of five runs with different random seeds, along with an unbiased estimate of the standard deviation. 

\subsection{Ablative Study}
Here an ablative study of the modeling decisions is conducted. Specifically, the effects of pretraining and class-balanced sampling are investigated. Results are presented in Tables \ref{tab:pretraining_cb_V} and \ref{tab:pretraining_cb_A}. Note that without pretraining or class balanced sampling the training degenerates into networks essentially predicting the BT label.

\parsection{Pretraining}
Here models pretrained on the larger dataset Kinetics-400 to models training from scratch on the proposed dataset are compared. It can be sen that for both modalities, pretraining gives a large boost in performance. 

\parsection{Class Balanced Sampling}
Class balanced sampling greatly increases performance. Furthermore, pretraining and class balanced sampling combined gives additive performance boosts, and hence are well used together.

\subsection{Late Fusion of Video and Audio}
\label{sec:late_fusion_resuls}
Given the predictions of the single modality models, the combined predictions are investigated. In Table \ref{tab:fusion} results are presented. It can be observed that the independence strategy outperforms the averaging strategy significantly. This follows the intuition that the audio modality may contain information which is difficult to ascertain from the visual modality, such as frightening sounds, and vice-versa. Furthermore the learned 2-stage training approach for fusion seems to give similar CAE scores as the independent version, however outperforming it in precision and recall. This indicates that the MLP model is more accurate, but that the errors it does make can be quite large. This additionally shows the importance of using an ordinal-aware metric.

\subsection{Model Probing Results}

The difference in average prediction between genders is calculated alongside the $p$-value for the genders being predicted differently.
This was performed with differently seeded models.
As can be seen in Table~\ref{tab:discrimination}, in three of the seeds the model has a significant bias toward videos with men being given a lower probability of belonging to the BT rating, while two do not.

While it is clear that trained models can certainly learn gender bias, it is also notable that in two of the cases it did not, showing that the model can be trained on the dataset without learning gender bias.

\begin{table}[h!]
    \caption{The average BT prediction for videos in the DeepFakeDetection dataset when the actor is a woman compared to when the actor is a man, along with the $p$-value for the hypothesis that there exists gender discrimination.}

    \centering
    \begin{tabular}{l|ccccc}
    \toprule
     &\multicolumn{5}{c}{Seed} \\

    & 1 & 2 & 3 & 4 & 5 \\
    \midrule
    
    Women &     0.853   & 0.950  & 0.893 & 0.882  & 0.936 \\
    Men &       0.801   & 0.934  & 0.897 & 0.855  & 0.925 \\
    \midrule
    $\Delta$ & 0.052 & 0.016 & 0.004 & 0.027 & 0.011 \\
    $p$-value & <0.001  & <0.001 & 0.614 & <0.001 & 0.116 \\
    \bottomrule
    \end{tabular}
    \label{tab:discrimination}
    \vspace{-8pt}
\end{table}



\section{Discussion and Conclusion}
In this paper VidHarm, a new multi-lingual professionally annotated dataset for detecting harmful content in video has been introduced. It was shown that compared to previous datasets it contains a greater diversity of content, and importantly is the only video dataset with professional labels. The properties of the dataset have been investigated and results show, among other things, that there is a strong correlation between the annotation of short clips vs the annotation for entire trailers. A novel metric for evaluation was proposed, and compared with previous metrics.
It was shown that training spatiotemporal and audio models on the proposed dataset yield impressive results when combining class balanced sampling, pretraining, and multimodal fusion. It was further shown that learned late fusion performed similarly to a fixed fusion model.
An investigation of potential gender biases in the trained models showed that while some models exhibit signs of bias, others did not.


\subsection{Ethical Considerations and Wider Implications}

Assessing whether content is harmful is a delicate issue and is a balancing act between freedom of information and protecting sensitive groups. Another issue is that of bias. In this work model biases have investigated in terms of gender, but further work is needed.

\section*{Acknowledgments}
\noindent
We thank Njutafilms/Studio S, and Folkets Bio.
This work has been supported by ELLIIT, the Strategic Area for ICT
research, funded by the Swedish Government, by Vinnova through
grant 2020-04057, and by Wallenberg Artificial Intelligence, Autonomous Systems and Software
Program (WASP) funded by Knut and Alice Wallenberg
Foundation.

{\small
\bibliographystyle{ieee_fullname}
\bibliography{references}
}
\clearpage




\end{document}